\documentclass[11pt,a4paper]{article}
\usepackage[hyperref]{acl2020}
\usepackage{times}
\usepackage{latexsym}
\usepackage{multirow}
\usepackage{multicol}
\usepackage{graphicx}
\usepackage{algorithm}
\usepackage{algorithmic}
\usepackage{amsmath}
\usepackage{booktabs}
\usepackage{url}

\usepackage{microtype}

\aclfinalcopy 

\title{Meta-KD: A Meta Knowledge Distillation Framework for Language Model Compression across Domains}

\date{}

\author{Haojie Pan$^{1,2}$\thanks{\ \ H. Pan and C. Wang contributed equally to this work.}, Chengyu Wang$^{2}$\footnotemark[1], Minghui Qiu$^{2}$\thanks{\ \ M. Qiu is the corresponding author.}, Yichang Zhang$^{2}$, Yaliang Li$^{2}$, Jun Huang$^{2}$\\
$^1$ Zhejiang Lab\\
$^2$ Alibaba Group\\
 {\tt \{haojie.phj, chengyu.wcy, minghui.qmh\}@alibaba-inc.com}\\
 {\tt \{yichang.zyc, yaliang.li, huangjun.hj\}@alibaba-inc.com}}

\begin{document}
\maketitle
\begin{abstract}
Pre-trained language models have been applied to various NLP tasks with considerable performance gains. However, the large model sizes, together with the long inference time, limit the deployment of such models in real-time applications. 
One line of model compression approaches considers knowledge distillation to distill large teacher models into small student models. Most of these studies focus on single-domain only, which ignores the transferable knowledge from other domains. We notice that training a teacher with transferable knowledge digested across domains can achieve better generalization capability to help knowledge distillation.
Hence we propose a Meta-Knowledge Distillation (Meta-KD) framework to build a meta-teacher model that captures transferable knowledge across domains and passes such knowledge to students. 
Specifically, we explicitly force the meta-teacher to capture transferable knowledge at both instance-level and feature-level from multiple domains, and then propose a meta-distillation algorithm to learn single-domain student models with guidance from the meta-teacher.
Experiments on public multi-domain NLP tasks show the effectiveness and superiority of the proposed Meta-KD framework. Further, we also demonstrate the capability of Meta-KD in the settings where the training data is scarce.
\end{abstract}

\section{Introduction} \label{sec:intro}


Pre-trained Language Models (PLM) such as BERT~\citep{conf/naacl/DevlinCLT19} and XLNet~\citep{conf/nips/YangDYCSL19} have achieved significant success with the two-stage ``pre-training and fine-tuning'' process.
Despite the performance gain achieved in various NLP tasks, the large number of model parameters and the long inference time have become the bottleneck for PLMs to be deployed in real-time applications, especially on mobile devices~\citep{DBLP:journals/corr/abs-1909-10351,DBLP:conf/acl/SunYSLYZ20,DBLP:journals/corr/abs-2006-11316}. 
Thus, there are increasing needs for PLMs to reduce the model size and the computational overhead while keeping the prediction accuracy. 

\begin{figure}[t]
\centering
\includegraphics[height=0.35\textwidth]{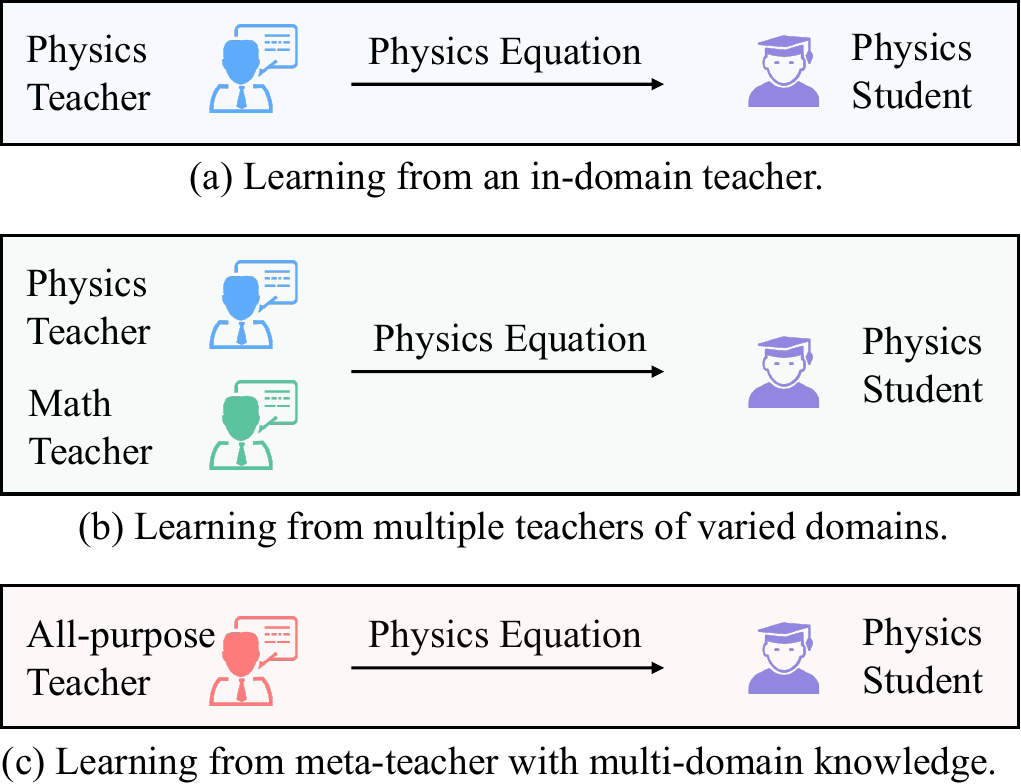}
\caption{A motivation example of academic learning. A physics student may learn physics equations better with a powerful all-purpose teacher.}\label{fig:intros}
\end{figure}

Knowledge Distillation (KD)~\cite{hiton_distill} is one of the promising ways to distill the knowledge from a large ``teacher'' model to a small ``student'' model. Recent studies show that KD can be applied to compress PLMs with acceptable performance loss~\citep{DBLP:journals/corr/abs-1910-01108,DBLP:journals/corr/abs-1908-09355,DBLP:journals/corr/abs-1909-10351,DBLP:journals/corr/abs-1908-08962,DBLP:conf/ijcai/ChenLQWLDDHLZ20}. 
However, those methods mainly focus on single-domain KD. Hence, student models can only learn from their in-domain teachers, paying little attention to acquiring knowledge from other domains.
It has been shown that it is beneficial to consider cross-domain information for KD, by either training a teacher using cross-domain data or multiple teachers from multiple domains~\citep{DBLP:conf/kdd/YouX0T17,DBLP:journals/corr/abs-1911-03588,DBLP:conf/wsdm/YangSGLJ20,DBLP:conf/aaai/PengJLCCZ20}. Consider an academic scenario in Figure \ref{fig:intros}. A typical way for a physics student to learn physics equations is to directly learn from his/her physics teacher. If we have a math teacher to teach him/her basic knowledge of equations, the student can obtain a better understanding of physics equations. This ``knowledge transfer'' technique in KD has been proved efficient only when two domains are close to each other~ \citep{DBLP:journals/corr/abs-1908-09122}. 
In reality, however, it is highly risky as teachers of other domains may pass non-transferable knowledge to the student model, which is irrelevant to the current domain and hence harms the overall performance~\citep{DBLP:conf/aaai/Tan0P017,DBLP:journals/corr/abs-2003-13003}.  Besides, current studies find multi-task fine-tuning of BERT does not necessarily yield better performance across all the tasks ~\citep{DBLP:conf/cncl/SunQXH19}.

To address these issues, we leverage the idea of meta-learning to capture transferable knowledge across domains, as recent studies have shown that meta-learning can improve the model generalization ability across domains~\citep{DBLP:conf/icml/FinnAL17,DBLP:conf/nips/JavedW19,DBLP:journals/corr/abs-2007-09604,DBLP:conf/acl/YeGCCXZWZC20}. 
We further notice that meta-knowledge is also helpful for cross-domain KD. 
Re-consider the example in Figure \ref{fig:intros}. If we have an ``all-purpose teacher'' (i.e.,~the meta-teacher) who has the knowledge of both physics principles and mathematical equations (i.e., the general knowledge of the two courses), the student may learn physics equations better with the teacher, compared to the other two cases. Hence, it is necessary to train an ``all-purpose teacher'' model for domain-specific student models to learn.



In this paper, we propose the Meta-Knowledge Distillation (Meta-KD) framework, which facilities cross-domain KD.
Generally speaking, Meta-KD consists of two parts, \textit{meta-teacher learning} and \textit{meta-distillation}. 
Different from the K-way N-shot problems addressed in traditional meta-learning~\cite{DBLP:journals/corr/abs-1810-03548}, we propose to train a ``meta-learner'' as the meta-teacher, which learns the transferable knowledge across domains so that it can fit new domains easily.
The meta-teacher is jointly trained with multi-domain datasets to 
acquire the~\emph{instance-level} and~\emph{feature-level} meta-knowledge. For each domain, the student model learns to solve the task over a domain-specific dataset with guidance from the meta-teacher. To improve the student's distillation ability, the meta-distillation module minimizes the distillation loss from both intermediate layers, output layers, and transferable knowledge, combined with domain-expertise weighting techniques.

To verify the effectiveness of Meta-KD, we conduct extensive experiments on two NLP tasks across multiple domains, namely natural language inference~\citep{MNLI_wiliam} and sentiment analysis ~\citep{DBLP:conf/acl/BlitzerDP07}. Experimental results show the effectiveness and superiority of the proposed Meta-KD framework. 
Moreover, we find our method performs well especially when i) the in-domain dataset is very small or ii) there is no in-domain dataset during the training of the meta-teacher.
In summary, the contributions of this study can be concluded as follows:
\begin{itemize}
\item To the best of our knowledge, this work is the first to explore the idea of meta-teacher learning for PLM compression across domains.
\item We propose the Meta-KD framework to address the task. In Meta-KD, the meta-teacher digests transferable knowledge across domains, and selectively passes the knowledge to student models with different domain expertise degrees.
\item We conduct extensive experiments to demonstrate the superiority of Meta-KD and also explore the capability of this framework in the settings where the training data is scarce. 
\end{itemize}


The rest of this paper is summarized as follows. Section~\ref{sec:related_works} describes the related work. The detailed techniques of the Meta-KD framework are presented in Section~\ref{sec:method}. The experiments are reported in Section~\ref{sec:experiments}. Finally, we conclude our work and discuss the future work in Section~\ref{sec:conlusion}.
\footnote{The experimental code can be found in \url{https://github.com/alibaba/EasyTransfer/tree/master/scripts/metaKD}.}

\section{Related Work} \label{sec:related_works}

Our study is close to the following three lines of studies, introduced below.

\subsection{Knowledge Distillation (KD)} 
KD was first proposed by~\cite{hiton_distill}, aiming to transfer knowledge from an ensemble or a large model into a smaller, distilled model. Most of the KD methods focus on utilizing either the dark knowledge, i.e., predicted outputs~\citep{hiton_distill,DBLP:conf/aaai/ChenMWF020,DBLP:conf/icml/FurlanelloLTIA18,DBLP:conf/kdd/YouX0T17} or hints, i.e., the intermediate representations ~\citep{DBLP:journals/corr/RomeroBKCGB14,DBLP:conf/cvpr/YimJBK17,DBLP:conf/kdd/YouX0T17} or the relations between layers ~\citep{DBLP:conf/cvpr/YimJBK17,DBLP:conf/nips/TarvainenV17} of teacher models.~\citet{DBLP:conf/kdd/YouX0T17} also find that multiple teacher networks together can provide comprehensive guidance that is beneficial for training the student network.~\citet{DBLP:journals/corr/RuderGB17} show that multiple expert teachers can improve the performances of sentiment analysis on unseen domains. \citet{DBLP:conf/iclr/TanRHQZL19} apply the multiple-teachers framework in KD to build a state-of-the-art multilingual machine translation system. \citet{l2a} considers to build a model to automatically augment data for KD. Our work is one of the first attempts to learn a meta-teacher model that digest transferable knowledge from multiple domains to benefit KD on the target domain.

\subsection{PLM Compression}
Due to the massive number of parameters in PLMs, it is highly necessary to compress PLMs for application deployment.
Previous approaches on compressing PLMs such as BERT ~\cite{conf/naacl/DevlinCLT19} include KD ~\citep{hiton_distill}, parameter sharing ~\citep{DBLP:conf/iclr/UllrichMW17}, pruning \citep{DBLP:conf/nips/HanPTD15} and quantization ~\citep{DBLP:journals/corr/GongLYB14}. In this work, we mainly focus on KD for PLMs. In the literature, \citet{DBLP:journals/corr/abs-1903-12136} distill BERT into BiLSTM networks to achieve comparable results with ELMo ~\citep{DBLP:conf/naacl/PetersNIGCLZ18}.
\citet{kd4tl} studies cross-domain KD to facilitate cross-domain knowledge transferring.
\citet{DBLP:journals/corr/abs-1909-11687} use dual distillation to reduce the vocabulary size and the embedding size. DistillBERT ~\citep{DBLP:journals/corr/abs-1910-01108} applies KD loss in the pre-training stage, while BERT-PKD~\citep{DBLP:journals/corr/abs-1908-09355} distill BERT into shallow Transformers in the fine-tuning stage. TinyBERT~\citep{DBLP:journals/corr/abs-1909-10351} further distills BERT with a two-stage KD process for hidden attention matrices and embedding matrices. AdaBERT ~\citep{DBLP:conf/ijcai/ChenLQWLDDHLZ20} uses neural architecture search to adaptively find small architectures.
Our work improves the prediction accuracy of compressed PLMs by leveraging cross-domain knowledge, which is complementary to previous works.

\subsection{Transfer Learning and Meta-learning}
TL has been proved to improve the performance on the target domain by leveraging knowledge from related source domains ~\citep{DBLP:journals/tkde/PanY10,DBLP:conf/emnlp/MouMYLX0J16,DBLP:conf/acl/LiuQH17,DBLP:conf/iclr/YangSC17}. In most NLP tasks, the ``shared-private'' architecture is applied to learn domain-specific representations and domain-invariant features ~\citep{DBLP:conf/emnlp/MouMYLX0J16,DBLP:conf/acl/LiuQH17,DBLP:conf/naacl/ChenYZLB18,DBLP:conf/www/ChenQYZHLB19}. Compared to TL, the goal of meta-learning is to train meta-learners that can adapt to a variety of different tasks with little training data ~\citep{DBLP:journals/corr/abs-1810-03548}. 
A majority of meta-learning methods for include metric-based~\cite{DBLP:conf/nips/SnellSZ17,DBLP:conf/cvpr/PanYLWNM19}, model-based~\cite{DBLP:conf/icml/SantoroBBWL16,DBLP:conf/iclr/BartunovROL20} and model-agnostic approaches~\cite{DBLP:conf/icml/FinnAL17,DBLP:conf/nips/FinnXL18,DBLP:conf/nips/VuorioSHL19}. Meta-learning can also be applied to KD in some computer vision tasks~\citep{DBLP:journals/corr/abs-1710-07535,DBLP:conf/icml/yunhun19,DBLP:conf/eccv/benliu20,DBLP:conf/aaai/BaiWKL20,DBLP:conf/cvpr/LiL0Z20}. For example, \citet{DBLP:journals/corr/abs-1710-07535} record per-layer meta-data for the teacher model to reconstruct a training set, and then adopts a standard training procedure to obtain the student model. 
In our work, we use instance-based and feature-based meta-knowledge across domains for the KD process.

\section{The Meta-KD Framework} \label{sec:method}

In this section, we formally introduce the Meta-KD framework. We begin with a brief overview of Meta-KD. After that, the techniques are elaborated.

\subsection{An Overview of Meta-KD}
Take text classification as an example. Assume there are $K$ training sets, corresponding to $K$ domains. In the $k$-th dataset $\mathcal{D}_k = \{ X_k^{(i)}, y_k^{(i)} \}_{i=1}^{N_k} $, $X_k^{(i)}$ is the $i$-th sample~\footnote{$X_k^{(i)}$ can be a sentence, a sentence pair or any other textual units, depending on the task inputs.} and $y_k^{(i)}$ is the class label of $X_k^{(i)}$. $N_k$ is the total number of samples in $\mathcal{D}_k$. Let $\mathcal{M}_k$ be the large PLM fine-tuned on $\mathcal{D}_k$. Given the $K$ datasets, the goal of Meta-KD is to obtain the $K$ student models $\mathcal{S}_1,\cdots, \mathcal{S}_K$ that are small in size but has similar performance compared to the $K$ large PLMs, i.e., $\mathcal{M}_1,\cdots, \mathcal{M}_K$.

In general, the Meta-KD framework can be divided into the following two stages:
\begin{itemize}
    \item\textbf{Meta-teacher Learning}: Learn a meta-teacher $\mathcal{M}$ over all domains $\bigcup\limits_{k=1}^{K}\mathcal{D}_k$. The model digests transferable knowledge from each domain and has better generalization while supervising domain-specific students.
    
    \item\textbf{Meta-distillation}: Learn $K$ in-domain students $\mathcal{S}_1,\cdots, \mathcal{S}_K$ that perform well in their respective domains, given only in-domain data $\mathcal{D}_k$ and the meta-teacher $\mathcal{M}$ as input.
\end{itemize}

During the learning process of the meta-teacher, we consider both~\emph{instance-level} and~\emph{feature-level} transferable knowledge.
Inspired by prototype-based meta-learning~\citep{DBLP:conf/nips/SnellSZ17,DBLP:conf/cvpr/PanYLWNM19}, the meta-teacher model should memorize more information about prototypes. Hence, we compute sample-wise~\emph{prototype scores} as the~\emph{instance-level} transferable knowledge. The loss of the meta-teacher is defined as the sum of classification loss across all $K$ domains with~\emph{prototype-based},~\emph{instance-specific weighting}. Besides, it also learns~\emph{feature-level transferable knowledge} by adding 
a~\emph{domain-adversarial loss} as an auxiliary loss.
By these steps, the meta-teacher is more generalized and digests transferable knowledge before supervising student models.

For meta-distillation, each sample is weighted by a~\emph{domain-expertise score} to address the meta-teacher's capability for this sample. The~\emph{transferable knowledge} is also learned for the students from the meta-teacher. The overall meta-distillation loss is a combination of the Mean Squared Error (MSE) loss from intermediate layers of both models~\citep{DBLP:journals/corr/abs-1908-09355,DBLP:journals/corr/abs-1909-10351}, the soft cross-entropy loss from output layers~\citep{hiton_distill}, and the~\emph{transferable knowledge distillation loss}, with~\emph{instance-specific domain-expertise weighting} applied. 


\begin{figure*}[t]
\centering
\includegraphics[width=\textwidth]{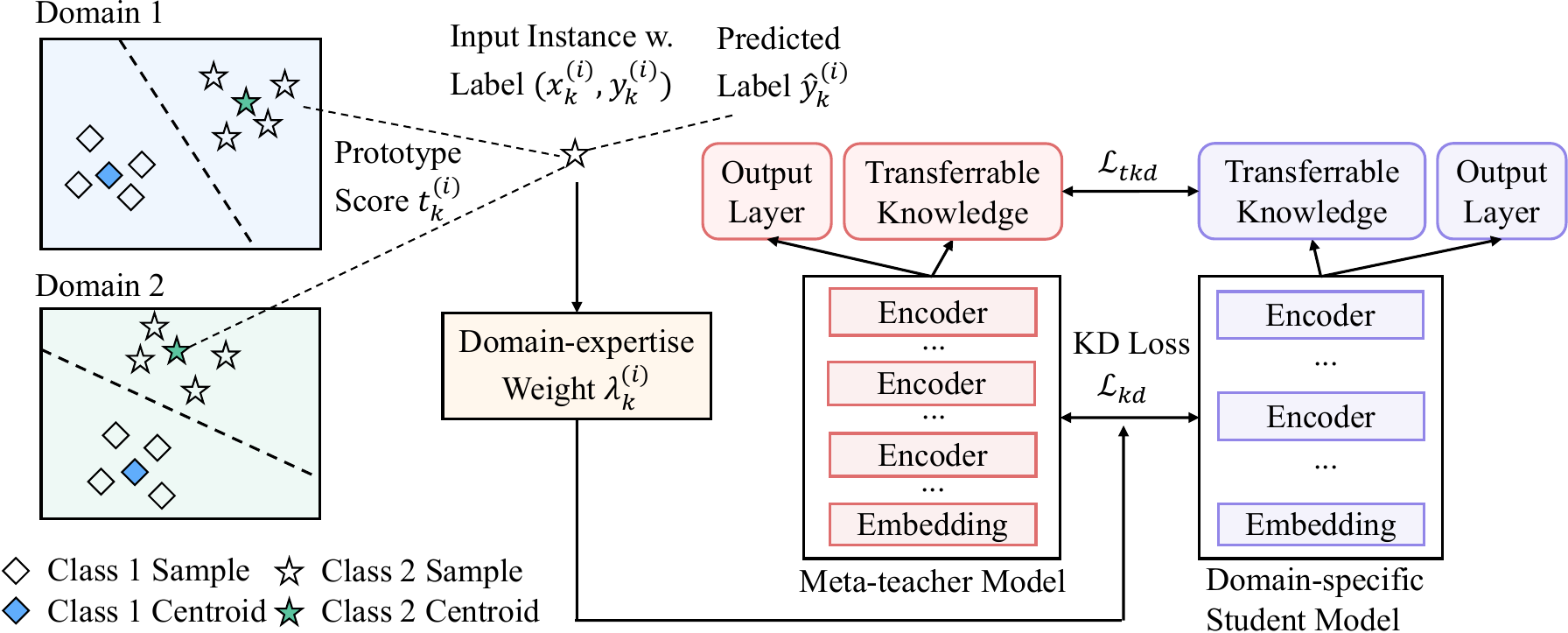}
\caption{An overview of meta-distillation and the neural architecture that we adopt for knowledge distillation.}\label{fig:meta_distill}
\end{figure*}

\subsection{Meta-teacher Learning} \label{subsec:meta_teacher}

We take BERT~\citep{conf/naacl/DevlinCLT19} as our base learner for text classification due to its wide popularity. For each sample $X_k^{(i)}$, the input is: \texttt{[CLS]}, $tok_{k,1}^{(i)}$, $tok_{k,2}^{(i)}$, $\cdots$, \texttt{[SEP]}, where $tok_{k,n}^{(i)}$ is the $n$-th token in $X_k^{(i)}$. The last hidden outputs of this sequence is denoted as $h_{[CLS]}, h(tok_{k,1}^{(i)}), h(tok_{k,2}^{(i)}), .., h(tok_{k,N}^{(i)})$, where $h(tok_{k,j}^{(i)})$ represents the last layer embedding of the $j$-th token in $X_k^{(i)}$, and $N$ is the maximum sequence length.
For simplicity, we define $h(X_k^{(i)})$ as the average pooling of the token embeddings, i.e., $h(X_k^{(i)})=\sum_{n=1}^{N}h(tok_{k,n}^{(i)})$.

\noindent \textbf{Learning Instance-level Transferable Knowledge.} 
To select~\emph{transferable instances across domains}, we compute a~\emph{prototype score} $t_k^{(i)}$ for each sample $X_k^{(i)}$. Here, we treat the~\emph{prototype representation} for the $m$-th class of the $k$-th domain:
$p_k^{(m)}=\frac{1}{\vert\mathcal{D}_k^{(m)}\vert }\sum_{X_k^{(i)}\in\mathcal{D}_k^{(m)}}h(X_k^{(i)})$,
where $\mathcal{D}_k^{(m)}$ is the $k$-th training set with the $m$-th class label. The~\emph{prototype score} $t_k^{(i)}$ is:
\begin{align*}
t_k^{(i)}= & \alpha\cos(p_k^{(m)},h(X_k^{(i)}))\\
& +\zeta\sum_{k'=1}^{K (k'\neq k)}\cos(p_{k'}^{(m)}, h(X_k^{(i)})),
\end{align*}
where $\cos$ is the cosine similarity function, $\alpha$ is a pre-defined hyper-parameter and $\zeta=\frac{1-\alpha}{K-1}$. We can see that the definition of the prototype score here is different from previous meta-learning, as we require that an instance $X_k^{(i)}$ should be close to its class prototype representation in the embedding space (i.e., $p_k^{(m)}$), as well as the prototype representations in out-of-domain datasets (i.e., $p_{k'}^{(m)}$ with $k'=1, \cdots, K, k'\neq k$). This is because the meta-teacher should learn more from instances that are~\emph{prototypical across domains} instead of~\emph{in-domain only}. For the text classification task, the cross-entropy loss of the meta-teacher is defined using the cross-entropy loss with the prototype score as a weight assigned to each instance.

\noindent \textbf{Learning Feature-level Transferable Knowledge.} 
Apart from the cross-entropy loss, we propose the~\emph{domain-adversarial loss} to increase the meta-teacher's ability for learning feature-level transferable knowledge.

For each sample $X_k^{(i)}$, we first learn an $\vert h(X_k^{(i)})\vert$-dimensional domain embedding of the true domain label $d_k^{(i)}$ by mapping one-hot domain representations to the embeddings, denoted as $\mathcal{E}_D (X_k^{(i)})$.
A sub-network is then constructed by:
\begin{equation*}
h_d(X_k^{(i)})) = \tanh((h(X_k^{(i)}) + \mathcal{E}_D(X_k^{(i)}))W + b),
\end{equation*}
where $W$ and $b$ are sub-network parameters. The domain-adversarial loss for $X_k^{(i)}$ is defined as:
\begin{equation*}
\mathcal{L}_{DA}(X_k^{(i)}) = -\sum_{k=1}^K \mathbf{1}_{k=z_k^{(i)}} \cdot\log \sigma(h_d(X_k^{(i)})),
\end{equation*}
where $\sigma$ is the $K$-way domain classifier, and $\mathbf{1}$ is the indicator function that returns 1 if $k=z_k^{(i)}$, and 0 otherwise. Here, $z_k^{(i)} \neq d_k^{(i)}$ is a false domain label of $X_k^{(i)}$\footnote{For ease of implementation, we shuffle the domain labels of all instances in a mini-batch.}. Hence, we deliberately maximize the probability that the meta-teacher makes the~\emph{wrong} predictions of domain labels. We call $h_d(X_k^{(i)}))$ as the~\emph{transferable knowledge} for $X_k^{(i)}$, which is more insensitive to domain differences. 

Let $\mathcal{L}_{CE}(X_k^{(i)})$ be the normal cross-entropy loss of the text classification task. The total loss of the meta-teacher $\mathcal{L}_{MT}$ is the combination of weighted $\mathcal{L}_{CE}(X_k^{(i)})$ and $\mathcal{L}_{DA}(X_k^{(i)})$, shown as follows:
\begin{align*}
\mathcal{L}_{MT} =\sum_{X_k^{(i)}\in \bigcup\limits_{k=1}^{K}\mathcal{D}_k} \frac{t_k^{(i)}\mathcal{L}_{CE}(X_k^{(i)}) + \gamma_1 \mathcal{L}_{DA}(X_k^{(i)})}{\sum_{k=1}^K\vert\mathcal{D}_k\vert},
\end{align*}
where $\gamma_1$ is the factor to represent how the domain-adversarial loss contributes to the overall loss.

\subsection{Meta-distillation} \label{subsec:meta_distillation}

We take BERT as our meta-teacher and use smaller BERT models as student models. The distillation framework is shown in Figure~\ref{fig:meta_distill}. In our work, we distill the knowledge in the meta-teacher model considering the following five elements: input embeddings, hidden states, attention matrices, output logits, and transferable knowledge.
The KD process of input embeddings, hidden states and attention matrices follows the common practice~\citep{DBLP:journals/corr/abs-1908-09355,DBLP:journals/corr/abs-1909-10351}. Recall that $\mathcal{M}$ and $\mathcal{S}_k$ are the meta-teacher and the $k$-th student model. Let $\mathcal{L}_{embd}(\mathcal{M},\mathcal{S}_k, X_k^{(i)})$, $\mathcal{L}_{hidn}(\mathcal{M},\mathcal{S}_k, X_k^{(i)})$ and $\mathcal{L}_{attn}(\mathcal{M},\mathcal{S}_k, X_k^{(i)})$ be the sample-wise MSE loss values of input embeddings, hidden states and attention matrices of the two models, respectively. Here, $\mathcal{L}_{embd}(\mathcal{M},\mathcal{S}_k, X_k^{(i)})$, $\mathcal{L}_{hidn}(\mathcal{M},\mathcal{S}_k, X_k^{(i)})$ and $\mathcal{L}_{attn}(\mathcal{M},\mathcal{S}_k, X_k^{(i)})$ refer to the sum of MSE values among multiple hidden layers. We refer interested readers to~\citet{DBLP:journals/corr/abs-1909-10351} for more details. $\mathcal{L}_{pred}(\mathcal{M},\mathcal{S}_k, X_k^{(i)})$ is the cross-entropy loss of ``softened'' output logits, parameterized by the temperature~\citep{hiton_distill}. A naive approach to formulating the total KD loss $\mathcal{L}_{kd}$ is the sum of all previous loss functions, i.e.,
\begin{align*}
\mathcal{L}_{kd}= \sum_{X_k^{(i)}\in \mathcal{D}_k} \big( \mathcal{L}_{embd}(\mathcal{M},\mathcal{S}_k, X_k^{(i)})+\\\mathcal{L}_{hidn}(\mathcal{M},\mathcal{S}_k, X_k^{(i)})+\mathcal{L}_{attn}(\mathcal{M},\mathcal{S}_k, X_k^{(i)})+\\ \mathcal{L}_{pred}(\mathcal{M},\mathcal{S}_k, X_k^{(i)}) \big).
\end{align*}

However, the above approach does not give special considerations to the transferable knowledge of the meta-teacher. Let $h_d^{\mathcal{M}}(X_k^{(i)})$ and $h_d^{\mathcal{S}}(X_k^{(i)})$ be the transferable knowledge of the meta-teacher and the student model w.r.t. the input $X_k^{(i)}$. We further define the ~\emph{transferable knowledge distillation loss} $\mathcal{L}_{TKD}(\mathcal{M},\mathcal{S}_k, X_k^{(i)})$ as follows:
\begin{align*}
& \mathcal{L}_{tkd}(\mathcal{M},\mathcal{S}_k, X_k^{(i)}) = \\
& \frac{1}{\vert\mathcal{D}_k\vert}\sum_{X_k^{(i)}\in \mathcal{D}_k} MSE \big( h_d^{\mathcal{M}}(X_k^{(i)}) W_{d}^{\mathcal{M}}, h_d^{\mathcal{S}}(X_k^{(i)}) \big) 
\end{align*}
where $W_{d}^{\mathcal{M}}$ is a learnable projection matrix to match the dimension between $h_d^{\mathcal{M}}(X_k^{(i)})$ and $h_d^{\mathcal{S}}(X_k^{(i)})$, and $MSE$ is the MSE loss function w.r.t. single element. In this way, we encourage student models to learn more transferable knowledge from the meta-teacher.

We further notice that although the knowledge of the meta-teacher should be highly transferable, there still exists the domain gap between the meta-teacher and domain-specific student models. In this work, for each sample $X_k^{(i)}$, we define the domain expertise weight $\lambda_k^{(i)}$ as follows:
\begin{align*}
\lambda_k^{(i)}= \frac{1 + t_k^{(i)}}{\exp^{(\hat{y}_k^{(i)} - y_k^{(i)})^2} + 1},
\end{align*}
where $\hat{y}_k^{(i)}$ is the predicted result of $X_k^{(i)}$'s class label. Here, the weight $\lambda_k^{(i)}$ is large when the meta-teacher model i) has a large prototype score $t_k^{(i)}$ and ii) makes correct predictions on the target input, i.e., $\hat{y}_k^{(i)}=y_k^{(i)}$. We can see that the weight reflects how well the meta-teacher can supervise the student on a specific input. Finally, we derive the complete formulation of the KD loss $\mathcal{L}_{kd}'$ as follows:
\begin{align*}
\mathcal{L}_{kd}'= \sum_{X_k^{(i)}\in\mathcal{D}_k} \lambda_k^{(i)} \big(\mathcal{L}_{embd}(\mathcal{M},\mathcal{S}_k, X_k^{(i)}) + \\
\mathcal{L}_{hidn}(\mathcal{M},\mathcal{S}_k, X_k^{(i)}) + 
\mathcal{L}_{attn}(\mathcal{M},\mathcal{S}_k, X_k^{(i)}) + \\
\mathcal{L}_{pred}(\mathcal{M},\mathcal{S}_k, X_k^{(i)}) \big)
+\gamma_2\mathcal{L}_{tkd}(\mathcal{M},\mathcal{S}_k, X_k^{(i)}) \big),
\end{align*}
where $\gamma_2$ is the \emph{transferable KD factor} to represent how the transferable knowledge distillation loss contributes to the overall loss.

\section{Experiments} \label{sec:experiments}

In this section, we conduct extensive experiments to evaluate the Meta-KD framework on two popular text mining tasks across domains.

\subsection{Tasks and Datasets}

We evaluate Meta-KD over natural language inference and sentiment analysis, using the following two datasets MNLI and Amazon Reviews. The data statistics are in Table \ref{table:dataset}.

\begin{itemize}
\item\textbf{MNLI}~\citep{MNLI_wiliam} is a large-scale, multi-domain natural language inference dataset for predicting the entailment relation between two sentences, containing five domains (genres). After filtering samples with no labels available, we use the original development set as our test set and randomly sample 10\% of the training data as a development set in our setting.

\item\textbf{Amazon Reviews}~\citep{DBLP:conf/acl/BlitzerDP07} is a multi-domain sentiment analysis dataset, widely used in multi-domain text classification tasks. The reviews are annotated as positive or negative. For each domain, there are 2,000 labeled reviews. We randomly split the data into train, development, and test sets. 
\end{itemize}

\begin{table}[!h] 
 \centering
 \begin{tabular}{lcccc} 
  \toprule 
  Dataset & Domain  & \#Train & \#Dev & \#Test  \\ 
  \midrule 
  \multirow{5}{*}{MNLI} & Fiction & 69,613 & 7,735 & 1,973 \\
   & Gov. & 69,615 & 7,735 & 1,945 \\
  & Slate & 69,575 & 7,731 & 1,955 \\
  & Telephone & 75,013 & 8,335 & 1,966 \\
  & Travel & 69,615 & 7,735 & 1,976 \\
  \midrule
  & Book & 1,631 & 170 & 199 \\
Amazon & DVD & 1,621 & 194 & 185 \\
Reviews & Elec. & 1,615 & 172 & 213 \\
  & Kitchen & 1,613 & 184 & 203 \\
 \bottomrule 
 \end{tabular} 
 \caption{
    Statistics of the two datasets.
 } 
\label{table:dataset}
\end{table}

\begin{table*}[!h] 
 \centering
 \begin{tabular}{lcccccc} 
  \toprule 
  Method  & Fiction & Government & Slate & Telephone & Travel & Average  \\ 
  \midrule 
$\text{BERT}_{\text{B}}\text{-single}$ & 82.2 & 84.2 & 76.7 & 82.4 & 84.2 & 81.9 \\
$\text{BERT}_{\text{B}}\text{-mix}$ & 84.8 & 87.2 & 80.5 & 83.8 & 85.5 & 84.4 \\
$\text{BERT}_{\text{B}}\text{-mtl}$ & 83.7 & 87.1 & 80.6 & 83.9 & 85.8 & 84.2 \\
Meta-teacher & 85.1 & 86.5 & 81.0 & 83.9 & 85.5 & 84.4 \\
  \midrule 
  
$\text{BERT}_{\text{B}}\text{-single}$ $\xrightarrow{\text{TinyBERT-KD}}$  $\text{BERT}_{\text{S}}$ & 78.8 & 83.2 & 73.6 & 78.8 & 81.9 & 79.3 \\

$\text{BERT}_{\text{B}}\text{-mix}$ $\xrightarrow{\text{TinyBERT-KD}}$  $\text{BERT}_{\text{S}}$ & 79.6 & 83.3 & 74.8 & 79.0 & 81.5 & 79.6 \\
$\text{BERT}_{\text{B}}\text{-mtl}$ $\xrightarrow{\text{TinyBERT-KD}}$  $\text{BERT}_{\text{S}}$ & 79.7 & 83.1 & 74.2 & 79.3 & 82.0 & 79.7 \\
$\text{Multi-teachers}$ $\xrightarrow{\text{MTN-KD}}$  $\text{BERT}_{\text{S}}$ & 77.4 & 81.1 & 72.2 & 77.2 & 78.0 & 77.2 \\
  \midrule 
$\text{Meta-teacher}$ $\xrightarrow{\text{TinyBERT-KD}}$ $\text{BERT}_{\text{S}}$ & 80.3 & 83.0 & \textbf{75.1} & 80.2 & 81.6 & 80.0 \\
$\text{Meta-teacher}$ $\xrightarrow{\text{Meta-distillation}}$ $\text{BERT}_{\text{S}}$ & \textbf{80.5} & \textbf{83.7} & 75.0 & \textbf{80.5} & \textbf{82.1} & \textbf{80.4} \\

 \bottomrule 
 \end{tabular}
 \caption{
    Results over MNLI (with five domains) in terms of accuracy (\%). Here $X \xrightarrow{\text{A}} Y$ means it uses $X$ as the teacher and $Y$ as the student, with $A$ as the KD method, hereinafter the same.
 } 
\label{table:results-mnli}
\end{table*}

\begin{table*}[!h] 
 \centering
 \begin{tabular}{lccccc} 
  \toprule 
  Method  & Books & DVD & Electronics & Kitchen & Average  \\ 
  \midrule 
$\text{BERT}_{\text{B}}\text{-single}$ & 87.9 & 83.8 & 89.2 & 90.6 & 87.9  \\
$\text{BERT}_{\text{B}}\text{-mix}$ & 89.9 & 85.9 & 90.1 & 92.1 & 89.5 \\
$\text{BERT}_{\text{B}}\text{-mtl}$ & 90.5 & 86.5 & 91.1 & 91.1 & 89.8 \\
Meta-teacher & 92.5 & 87.0 & 91.1 & 89.2 & 89.9 \\
 \midrule
$\text{BERT}_{\text{B}}\text{-single}$ $\xrightarrow{\text{TinyBERT-KD}}$  $\text{BERT}_{\text{S}}$ & 83.4 & 83.2 & 89.2 & 91.1 & 86.7 \\
$\text{BERT}_{\text{B}}\text{-mix}$ $\xrightarrow{\text{TinyBERT-KD}}$  $\text{BERT}_{\text{S}}$ & 88.4 & 81.6 & 89.7 & 89.7 & 87.3 \\
$\text{BERT}_{\text{B}}\text{-mtl}$ $\xrightarrow{\text{TinyBERT-KD}}$  $\text{BERT}_{\text{S}}$ & 90.5 & 81.6 & 88.7 & 90.1 & 87.7 \\
$\text{Multi-teachers}$ $\xrightarrow{\text{MTN-KD}}$  $\text{BERT}_{\text{S}}$ & 83.9 & 78.4 & 88.7 & 87.7 & 84.7 \\
 \midrule
$\text{Meta-teacher}$ $\xrightarrow{\text{TinyBERT-KD}}$ $\text{BERT}_{\text{S}}$ & 89.9 & 84.3 & 87.3 & \textbf{91.6} & 88.3 \\
$\text{Meta-teacher}$ $\xrightarrow{\text{Meta-distillation}}$ $\text{BERT}_{\text{S}}$ & \textbf{91.5} & \textbf{86.5} & \textbf{90.1} & 89.7 & \textbf{89.4} \\
 \bottomrule 
 \end{tabular}
 \caption{
    Results over Amazon reviews (with four domains) in terms of accuracy (\%).
 } 
\label{table:results-amazon}
\end{table*}

\subsection{Baselines}
For the teacher side, to evaluate the cross-domain distillation power of the meta-teacher model, we consider the following models as baseline teachers:
\begin{itemize}
\item \textbf{BERT-single}: Train the BERT teacher model on the target distillation domain only. If we have $K$ domains, then we will have $K$ BERT-single teachers.
\item \textbf{BERT-mix}: Train the BERT teacher on a combination of $K$-domain datasets. Hence, we have one BERT-mix model as the teacher model for all domains.
\item \textbf{BERT-mtl}: Similar to the ``one-teacher'' paradigm as in BERT-mix, but the teacher model is generated by the multi-task fine-tuning approach~\citep{DBLP:conf/cncl/SunQXH19}.
\item \textbf{Multi-teachers}: It uses $K$ domain-specific BERT-single models to supervise $K$ student models, ignoring the domain difference.
\end{itemize}

For the student side, we follow TinyBERT ~\citep{DBLP:journals/corr/abs-1909-10351} to use smaller BERT models as our student models. In single-teacher baselines (i.e., BERT-single, BERT-mix and BERT-mtl), we use  TinyBERT-KD as our baseline KD approach. In multi-teachers, because TinyBERT-KD does not naturally support distilling from multiple teacher models, we implement a variant of the TinyBERT-KD process based on MTN-KD~\citep{DBLP:conf/kdd/YouX0T17}, which uses averaged softened outputs as the incorporation of multiple teacher networks in the output layer. In practice, we first learn the representations of the student models by TinyBERT, then apply MTN-KD for output-layer KD.

\subsection{Implementation Details}

In the implementation, we use the original $\textbf{BERT}_{\textbf{B}}$ model (L=12, H=768, A=12, Total Parameters=110M) as the initialization of all of the teachers, and use the $\textbf{BERT}_{\textbf{S}}$ model (L=4, H=312, A=12, Total Parameters=14.5M) as the initialization of all the students\footnote{\url{https://github.com/huawei-noah/Pretrained-Language-Model/tree/master/TinyBERT}}.

The hyper-parameter settings of the meta-teacher model are as follows. We train 3-4 epochs with the learning rate to be 2e-5. The batch size and $\gamma_1$ are chosen from \{16, 32, 48\} and \{0.1, 0.2, 0.5\}, respectively. All the hyper-parameters are tuned on the development sets.

For meta-distillation, we choose the hidden layers in \{3, 6, 9, 12\} of the teacher models in the baselines and the meta-teacher model in our approach to learn the representations of the student models. Due to domain difference, we train student models in 3-10 epochs, with a learning rate of 5e-5. The batch size and $\gamma_2$ are tuned from \{32, 256\} and \{0.1, 0.2, 0.3, 0.4, 0.5\} for intermediate-layer distillation, respectively. Following~\citet{DBLP:journals/corr/abs-1909-10351}, for prediction-layer distillation, we run the method for 3 epochs, with the batch size and learning rate to be 32 and 3e-5. The experiments are implemented on PyTorch and run on 8 Tsela V100 GPUs.

\subsection{Experimental Results}

Table~\ref{table:results-mnli} and Table~\ref{table:results-amazon} show the general testing performance over MNLI and Amazon Reviews of baselines and Meta-KD, in terms of accuracy. From the results, we have the following three major insights:
\begin{itemize}
    \item Compared to all the baseline teacher models, using the meta-teacher for KD consistently achieves the highest accuracy in both datasets. Our method can help to significantly reduce model size while preserving similar performance, especially in Amazon review, we reduce the model size to 7.5x smaller with only a minor performance drop (from 89.9 to 89.4). 
    \item The meta-teacher has similar performance as BERT-mix and BERT-mtl, but shows to be a better teacher for distillation, as  $\text{Meta-teacher}$ $\xrightarrow{\text{TinyBERT-KD}}$ $\text{BERT}_{\text{S}}$  and  $\text{Meta-teacher}$ $\xrightarrow{\text{Meta-distillation}}$ $\text{BERT}_{\text{S}}$  have better performance than other methods. This shows the meta-teacher is capable of learning more transferable knowledge to help the student. The fact that Meta-teacher $\rightarrow$ Meta-distillation has better performance than other distillation methods confirms the effectiveness of the proposed Meta-KD method. 
    \item Meta-KD gains more improvement on the small datasets than large ones, e.g. it improves from 86.7 to 89.4 in Amazon Reviews while 79.3 to 80.4 in MNLI. This motivates us to explore our model performance on domains with few or no training samples
\end{itemize}


\subsection{Ablation Study}

We further investigate Meta-KD's capability with regards to different portion training data for both of two phases and explore how the transferable knowledge distillation loss contributes to final results.

\begin{table}[!t] 
 \small
 \centering
 \begin{tabular}{llc} 
 \toprule
 \multicolumn{2}{l}{Method} & Accuracy \\
 \midrule  
 \multicolumn{2}{l}{$\text{BERT}_{\text{B}}$-s (fiction)} & 82.2\% \\
 \multicolumn{2}{l}{Meta-teacher (w/o fiction)} & 81.6\% \\ 
 \midrule
 \multicolumn{2}{l}{$\text{BERT}_{\text{B}}$-s (fiction) $\xrightarrow{\text{TinyBERT-KD}}$ $\text{BERT}_{\text{S}}$} & 78.8\% \\
 \midrule
 \multicolumn{2}{l}{$\text{BERT}_{\text{B}}$-s (govern) $\xrightarrow{\text{TinyBERT-KD}}$ $\text{BERT}_{\text{S}}$} & 75.3\% \\
 \multicolumn{2}{l}{$\text{BERT}_{\text{B}}$-s (telephone) $\xrightarrow{\text{TinyBERT-KD}}$ $\text{BERT}_{\text{S}}$} & 75.6\% \\
 \multicolumn{2}{l}{$\text{BERT}_{\text{B}}$-s (slate) $\xrightarrow{\text{TinyBERT-KD}}$ $\text{BERT}_{\text{S}}$} & 77.1\% \\
 \multicolumn{2}{l}{$\text{BERT}_{\text{B}}$-s (travel) $\xrightarrow{\text{TinyBERT-KD}}$ $\text{BERT}_{\text{S}}$} & 74.1\% \\
 \midrule
 \multicolumn{2}{l}{Meta-teacher $\xrightarrow{\text{TinyBERT-KD}}$ $\text{BERT}_{\text{S}}$} & \textbf{78.2\%} \\
 \bottomrule
 \end{tabular}
\caption{Results under the setting where no in-domain data used for meta-teacher learning on MNLI. Here, ``$\text{BERT}_{\text{B}}$-s'' refers to the ``$\text{BERT}_{\text{B}}$-single'' method. The distillation is performed on the ``fiction'' domain data. We report accuracy on the domain dataset.}\label{table:zero-shot}
\end{table}

\subsubsection{No In-domain Data during Meta-teacher Learning} 
In this set of experiments, we consider a special case where we assume all the ``fiction'' domain data in MNLI is unavailable. Here, we train a meta-teacher without the ``fiction'' domain dataset and use the distillation method proposed in ~\citet{DBLP:journals/corr/abs-1909-10351} to produce the student model for the ``fiction'' domain with in-domain data during distillation. The results are shown in Table~\ref{table:zero-shot}. We find that KD from the meta-teacher can have large improvement, compared to KD from other out-domain teachers. Additionally, learning from the out-domain meta-teacher has a similar performance to KD from the in-domain ``fiction'' teacher model itself. It shows the Meta-KD framework can be applied in applications for emerging domains.

\begin{figure}[t]
\centering
\includegraphics[width=0.45\textwidth]{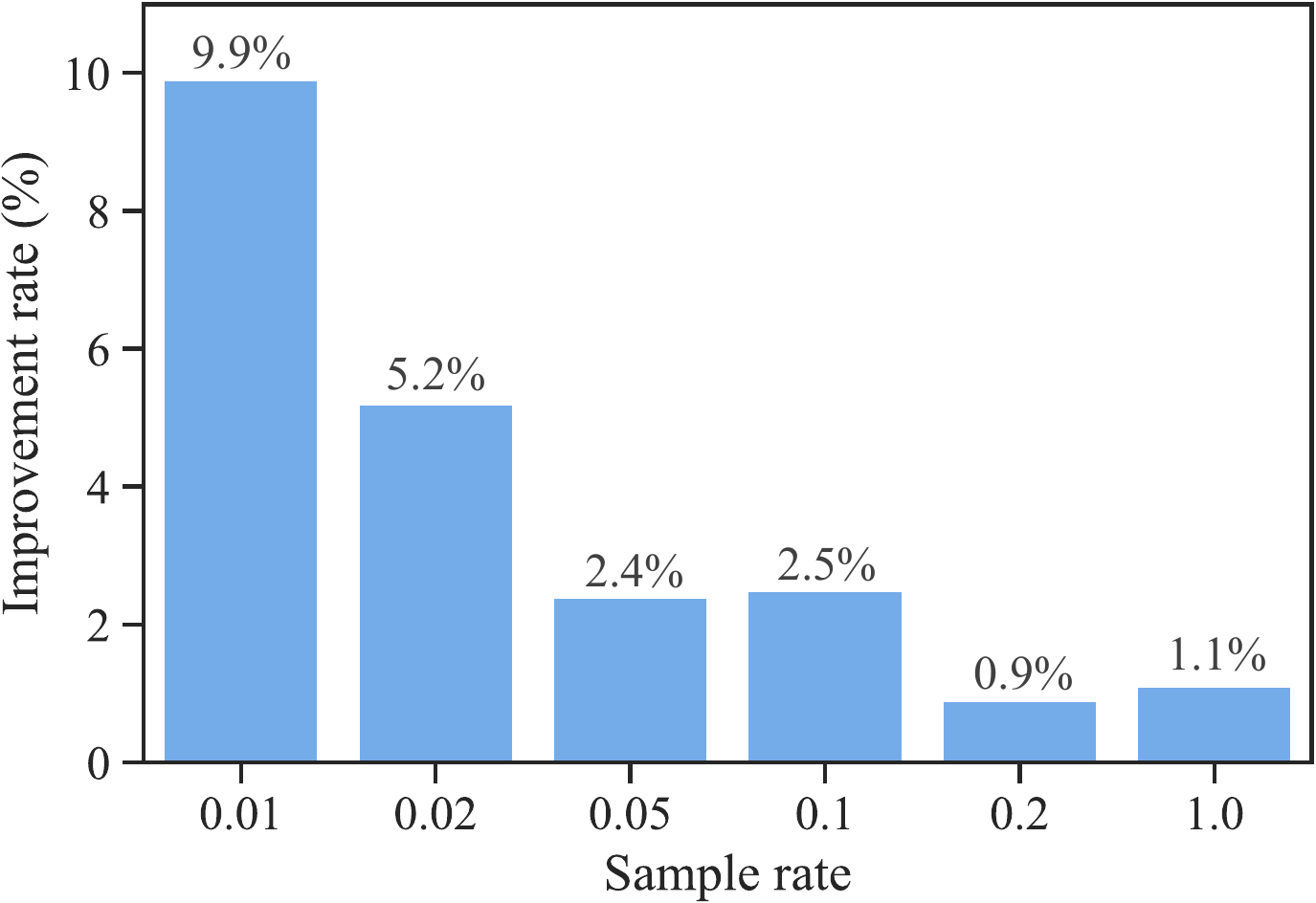}
\caption{Improvement rate w.r.t different portion (sample rate) of training data in usage.}\label{fig:few-shot}
\end{figure}

\subsubsection{Few In-domain Data Available during Meta-distillation}
We randomly sample a part of the MNLI dataset as the training data in this setting. The sample rates that we choose include 0.01, 0.02, 0.05, 0.1 and 0.2. The sampled domain datasets are employed for training student models when learning from the in-domain teacher or the meta-teacher. The experimental results are shown in Figure \ref{fig:few-shot}, with results reported by the improvement rate in averaged accuracy. The experimental results show that when less data is available, the improvement rate is much larger. For example, when we only have 1\% of the original MNLI training data, the accuracy can be increased by approximately 10\% when the student tries to learn from the meta-teacher. It shows Meta-KD can be more beneficial when we have fewer in-domain data.

\subsubsection{Influence of the Transferable Knowledge Distillation Loss}
\begin{figure}[t]
\centering
\includegraphics[width=0.45\textwidth]{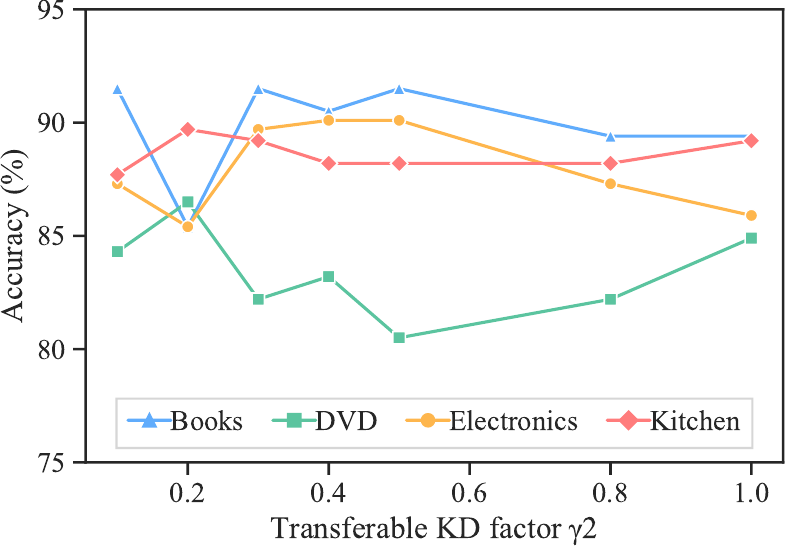}
\caption{Model performance w.r.t. the transferable KD factor $\gamma_2$}\label{fig:gamma2-tuning}
\end{figure}

Here, we explore how the \emph{transferable KD factor} $\gamma_2$ affects the distillation performance over the Amazon Reviews dataset. We tune the value of $\gamma_2$ from 0.1 to 1.0, with results are shown in Figure \ref{fig:gamma2-tuning}. We find that the optimal value of $\gamma_2$ generally lies in the range of 0.2 - 0.5. The trend of accuracy is different in the domain ``DVD'' is different from those of the remaining three domains. This means the benefits from transferable knowledge of the meta-teacher vary across domains.

\section{Conclusion and Future Work} \label{sec:conlusion}
In this work, we propose the Meta-KD framework which consists of meta-teacher learning and meta distillation to distill PLMs across domains. Experiments on two widely-adopted public multi-domain datasets show that Meta-KD can train a meta-teacher to digest knowledge across domains to help better teach in-domain students.
Quantitative evaluations confirm the effectiveness of Meta-KD and also show the capability of Meta-KD in the settings where the training data is scarce i.e. there is no or few in-domain data. 
In the future, we will examine the generalization capability of Meta-KD in other application scenarios and apply other meta-learning techniques to KD for PLMs.

\section*{Acknowledgements}
This work is supported by Open Research Projects of Zhejiang Lab (No. 2019KD0AD01/004). Any opinions, findings, and conclusions or recommendations expressed in this material are those of the authors and do not necessarily reflect those of the sponsor.

\bibliography{acl2020}
\bibliographystyle{acl_natbib}

\end{document}